\pdfoutput=1
\documentclass[11pt]{article}

\usepackage[final]{acl}

\usepackage{times}
\usepackage{latexsym}
\usepackage{amsmath}
\usepackage{cuted}
\usepackage{subcaption}
\usepackage{physics}
\usepackage{footmisc}
\usepackage{algorithm}[1]
\usepackage{algpseudocode}
\usepackage[T1]{fontenc}
\usepackage[utf8]{inputenc}

\usepackage{microtype}

\usepackage{inconsolata}

\usepackage{graphicx}

\title{Tailored Conversations beyond LLMs: A RL-Based Dialogue Manager %for Adaptive and Phase aware Hybrid Motivational Interviewing Systems 
}
\author{Lucie Galland \\
  ISIR \\ Sorbonne University \\ Paris, France \\
  \texttt{galland@isir.upmc.fr} \\\And
  Catherine Pelachaud \\
  CNRS/ISIR \\ Sorbonne University \\ Paris, France \\
  \texttt{pelachaud@isir.upmc.fr} \\\And
  Florian Pecune \\
  CNRS/SANPSY \\ Univ. Bordeaux\\ Bordeaux, France \\
  \texttt{pecune@u-bordeaux.fr} \\}

\begin{document}
\maketitle
\begin{abstract}
In this work, we propose a novel framework that integrates large language models (LLMs) with an RL-based dialogue manager for open-ended dialogue with a specific goal. By leveraging hierarchical reinforcement learning to model the structured phases of dialogue and employ meta-learning to enhance adaptability across diverse user profiles, our approach enhances adaptability and efficiency, enabling the system to learn from limited data, transition fluidly between dialogue phases, and personalize responses to heterogeneous patient needs. We apply our framework to Motivational Interviews, aiming to foster behavior change, and demonstrate that the proposed dialogue manager outperforms a state-of-the-art LLM baseline in terms of reward, showing a potential benefit of conditioning LLMs to create open-ended dialogue systems with specific goals.
\end{abstract}

\section{Introduction}

In recent years, the demand for mental health services has surged, outpacing the availability of resources and creating a substantial gap in access to care \cite{cameron2017towards}. As a result, many patients face extended waiting periods before receiving therapy \cite{cameron2017towards,denecke2020mental}. To address this challenge, virtual agents capable of simulating Motivational Interviewing (MI) have been proposed as a potential interim support system for individuals awaiting treatment. These agents can provide immediate assistance, particularly beneficial in therapeutic approaches requiring multiple sessions \cite{fiske2019your}. However, their role is not to replace human therapists but to serve as a supplementary tool that enhances existing therapeutic interventions. Motivational Interviewing (MI) poses a particularly complex challenge for dialogue systems, traditionally addressed through intricate rule-based frameworks \cite{prochaska2021therapeutic,olafsson2020towards}. However, recent advances in natural language processing (NLP) have paved the way for leveraging large language models (LLMs) such as GPT-like architectures \cite{baktash2023gpt} in such applications \cite{steenstra2024virtual}, significantly expanding the scope of dialogue systems across various domains. Although these models exhibit remarkable language generation capabilities, they also present significant limitations, many of which can be addressed through insights from "traditional" dialogue research. In particular, LLMs often lack the controllability and structured decision-making of conventional rule-based systems, which are more predictable and interpretable \cite{shidara2020analysis}. Rule-based domain-specific dialogue systems \cite{hadi2024large} offer notable advantages, including improved controllability, explainability, and the ability to integrate expert knowledge. However, they are typically less adaptable and more resource-intensive to develop. In contrast, LLMs demonstrate strong adaptability across domains but pose challenges in achieving control, transparency, and efficiency. Additionally, incorporating expert knowledge into LLMs often requires extensive domain-specific data \cite{hadi2024large}.  
Notably, reinforcement learning (RL)-based dialogue managers \cite{pecune2020framework} have shown promise in enhancing control and coherence in dialogue systems. 
Hence, a promising approach involves hybrid models that combine the strengths of both paradigms—leveraging the adaptability and generative capabilities of LLMs while integrating a domain-specific dialogue manager to regulate interactions \cite{abu2024supporting,galland2024simulating}. 

In this work, we investigate a hybrid approach in which an RL-based dialogue manager governs an LLM to simulate MI dialogues, aiming to balance adaptability and control for more effective virtual therapy support.

\section{Theoretical background: Motivational Interviewing (MI)}
\label{sec:context}
Motivational Interviewing (MI) is a therapeutic approach that emphasizes collaboration and supports behavioral change by guiding patients to explore the reasons and motivations behind their unhealthy behaviors. 
% \subsection{Dialogue with Multiple Phases}
% \begin{figure}[h]
% \centering
% \includegraphics[width=\linewidth]{latex/images/MI dialog phase.png}
% \caption{Phases of Motivational Interviewing}
% \label{fig:mi_phases}
% \end{figure}
Complex dialogues, such as those in Motivational Interviewing (MI), evolve through distinct phases, each guided by unique long-term strategies \cite{miller2012motivational}. %(see Figure \ref{fig:mi_phases}). 
The dialogue usually begins with an \textit{engaging} phase, where rapport is established, and patient engagement with the therapist is fostered. This is followed by a \textit{focusing} phase, where core issues, their underlying causes, and the patient’s background are identified to set a clear focus for the conversation. The third phase is the \textit{evoking } phase, which involves encouraging the patient’s motivation for change by eliciting and amplifying “change talk”. Finally, \textit{planning }involves developing a specific, actionable plan for behavior change based on the patient’s motivation and goals. Therapists must ensure that specific objectives, such as achieving high levels of engagement, clarifying core issues, and cultivating sufficient motivation, are met before transitioning between phases. Furthermore, the process is not strictly linear, as therapists may need to revisit earlier phases depending on the patients’ evolving motivation and engagement. Individual variability in engagement and motivation necessitates a flexible approach. For a virtual therapist employing MI, effectively navigating across these phases is crucial. This requires discerning when to progress to the next phase, when to revisit earlier phases, and how to adapt the interaction to align with each patient’s unique needs and circumstances.

\subsection{Patients' Profiles in MI}

Patients participating in motivational interviewing (MI) exhibit varying levels of readiness to change their behaviors. As proposed by \cite{galland2024emmiannonymous}, these patients can be classified into three categories: \textit{Open-to-Change}, \textit{Resistant-to-Change}, and \textit{Receptive}. \textit{Open-to-Change} individuals demonstrate a strong willingness to modify unhealthy behaviors. \textit{Resistant-to-Change} patients are generally reluctant to alter their current behaviors, showing a preference for maintaining the status quo. \textit{Receptive} patients, while initially exhibiting low motivation to change, gradually develop a higher motivation to adopt healthier behaviors as the conversation progresses. These classifications capture variations in patients' responses and therapists' strategies, as discussed in \cite{galland2024emmiannonymous}. The ability to adapt the flow of dialogue to these three patient profiles can significantly enhance the efficiency of the therapist's dialogue model. A dialogue system for MI should be able to take into account the particular challenges that such dialogues pose. Such a dialogue system should be able to navigate across phases while being able to adapt to different profiles of users. 

\section{Related Work}

Motivational Interviewing (MI) presents significant challenges for dialogue systems, as it necessitates both a structured progression through its four distinct phases —engagement, focusing, evoking, and planning \cite{miller2012motivational} — and adaptability to diverse patients' profiles \cite{galland2024emmiannonymous}. While existing systems such as Woebot \cite{prochaska2021therapeutic} have demonstrated the feasibility of MI-based chatbots by incorporating therapeutic frameworks like cognitive behavioral therapy (CBT) and mindfulness, they predominantly rely on static or rule-based architectures. Steenstral et al. \cite{steenstra2024virtual} identified the limitations of rule-based approaches in maintaining adherence to therapeutic protocols and proposed leveraging LLMs for this application, demonstrating promising results.
% However, these systems lack the ability to dynamically transition between MI phases or personalize interventions based on diverse patient profiles.  
The advent of large language models (LLMs) has transformed dialogue generation, offering new possibilities for MI-based interactions \cite{steenstra2024virtual}. Models such as GPT-like systems \cite{baktash2023gpt} exhibit strong generative capabilities and adaptability across diverse applications, which can be leveraged to enhance the social dimensions of interaction. This aspect is particularly relevant, as Kanaoka et al. \cite{kanaoka2015designing} emphasized the critical role of social engagement and rapport in facilitating behavioral change. Recent research has underscored the importance of cognitive modeling and adaptability in dialogue systems to more accurately account for the mental states of both the agent and the user. For instance, He et al. \cite{he2024planning} introduced dual reasoning mechanisms that enable LLMs to incorporate contextual nuances, while Zhang et al. \cite{zhang2020learning} explored interactive agent representations to improve dialogue coherence. Despite these advancements, LLMs remain constrained in terms of controllability and domain specificity \cite{shidara2020analysis}. Recent efforts have sought to enhance control and applicability, particularly within task-oriented dialogue systems. Yao et al. \cite{yao2023retroformer} employed reinforcement learning (RL) to optimize LLM prompting strategies; however, black-box LLMs continue to exhibit limitations in controllability and often generate repetitive responses. To address these shortcomings, Xu et al. \cite{xu2023small} demonstrated the advantages of integrating fine-tuned, smaller language models with larger LLMs, while Yu et al. \cite{yu2023prompt} employed Monte Carlo tree search for optimal action selection, improving both coherence and practical utility. Although these studies mark significant progress in hybrid model design, they often fall short of accommodating a broad spectrum of user profiles. Integrating the structured controllability of classical dialogue models with the generative flexibility of LLMs within an adaptive hybrid framework could facilitate more dynamic, personalized, and effective MI interactions. Reinforcement learning has been instrumental in optimizing dialogue policies and refining system behavior. Traditional RL-based approaches have primarily focused on enhancing user engagement and task success rates. Walker et al. \cite{walker2000application} and Li et al. \cite{li2016deep} demonstrated the efficacy of RL in training conversational agents for goal-oriented tasks, while Weber et al. \cite{weber2018shape} illustrated its effectiveness in selecting contextually appropriate actions, such as humor or sound effects, to enrich user experiences. More advanced frameworks have incorporated both social and task-oriented rewards \cite{pecune2020framework} or jointly trained user and dialogue policies \cite{takanobu2020multi}. Although these methods enable adaptation to different user profiles, they remain largely confined to task-oriented systems that rely on predefined natural language templates. In this work, we propose a novel framework that integrates LLMs with an RL-based dialogue manager to structure MI dialogues across different phases while dynamically adapting to diverse patient profiles. By synergizing structured control with generative flexibility, our approach enhances adaptability and efficiency, enabling the system to learn from limited data, transit fluidly between MI phases, and personalize responses to heterogeneous patient needs. The remainder of this paper is organized as follows: Section \ref{sec:method} details our proposed methodology, while Section \ref{sec:evaluation} describes our evaluation environment. Section \ref{sec:results} presents the experimental results, and Section \ref{sec:interpretation} provides an analysis and interpretation of our findings.  

\section{Method}
\label{sec:method}
\begin{figure}[h]
\centering
\includegraphics[width=\linewidth]{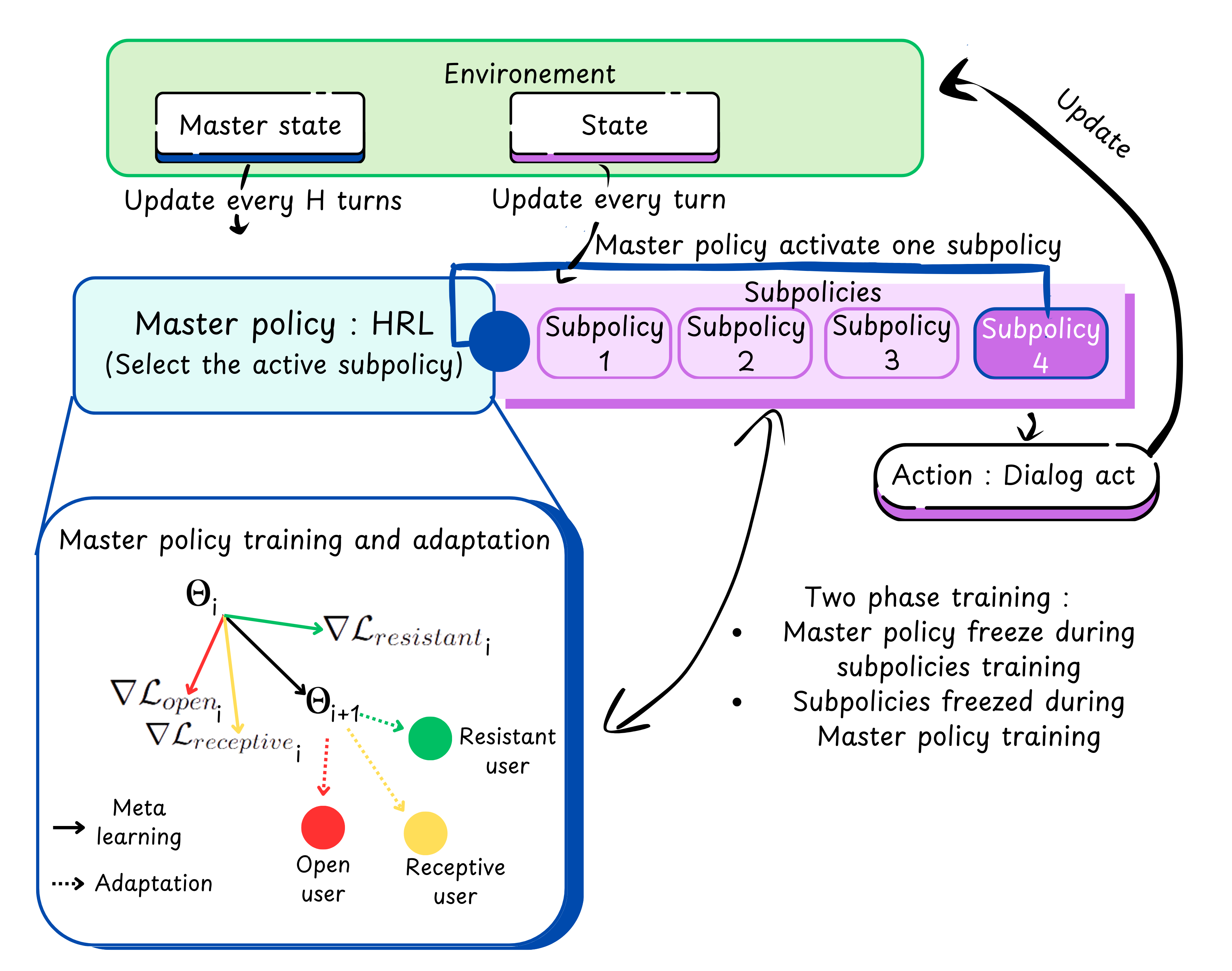}
\caption{Hierarchical architecture of the dialogue manager}
\label{fig:archi}
\end{figure}

This section outlines the methodology for developing a dialogue manager capable of navigating the distinct phases of Motivational Interviewing (MI) while adapting to diverse patients profile. The complete architecture is depicted in Figure \ref{fig:archi}.
\subsection{Problem description}
The objective of the proposed model is to predict the optimal action $a_t$ at each time step $t$, given the dialogue context $c$. Each action corresponds to a dialogue act representing the virtual therapist’s strategic behavior. The agent sentence is then generated by a conditioned large language model (LLM) that produces an utterance coherent with the context and realizes the selected dialogue act, as validated in \cite{galland2024simulating}.

\subsection{Hierarchical RL for Managing Dialogue Phases}
The dialogue manager employs a hierarchical reinforcement learning (HRL) framework to manage dialogue phases comprising a master policy and $N$ sub-policies. Each sub-policy is dedicated to a specific phase of the Motivational Interviewing (MI) process, while the master policy governs transitions between these phases over a fixed horizon $H$ (see Figure \ref{fig:archi}). The master policy orchestrates the dialogue by selecting the appropriate phase for the next $H$ dialogue turns based on the current master state. Meanwhile, sub-policies execute phase-specific strategies and actions, leveraging their respective sub-policy states. Notably, the master state and the sub-policy state differ in composition, as long-term planning requires distinct information from short-term decision-making. This hierarchical structure ensures dynamic and context-sensitive dialogue management, allowing real-time adjustments to both the patient's evolving needs and the interaction context. It balances global objectives, such as increasing motivation for behavior change, with more localized goals, such as answering patient inquiries. The fixed decision interval of $H$ turns reduces the training horizon, simplifying the learning process and enhancing the model’s adaptability to diverse users by focusing on shorter-term adjustments. By leveraging hierarchical reinforcement learning (HRL), the model effectively manages the different phases of MI dialogue while minimizing adaptation complexity. The sub-policies handle local objectives, while the master policy adjusts to user-specific global goals. Once the sub-policies are trained, adapting to a new user requires fine-tuning only the master policy, which operates on a smaller horizon and action space. This approach enables efficient transitions between MI phases, ensuring that interactions remain tailored to the needs of each individual user.

\subsection{Meta-Learning for User Adaptation}
To facilitate rapid adaptation to new users, the master policy is trained using meta-learning techniques—specifically, the Model-Agnostic Meta-Learning (MAML) algorithm \cite{finn2017model}. This approach allows the dialogue manager to generalize effectively across diverse user profiles while maintaining the flexibility to adapt quickly to novel ones. By leveraging MAML, the master policy learns shared strategies across the profiles of various patients and can fine-tune its behavior to accommodate individual patient needs, thus ensuring both robustness and personalization. In practice, after training, this meta-learned master policy serves as a well-informed initialization that approximates its behavior by averaging it across the profiles of users that have been seen. When deployed with human users for repeated interaction, it enables rapid personalization, requiring only a few interactions to adapt to the specific preferences and characteristics of a new user.

\subsection{Algorithm and Training Framework}
In this subsection, we present formally our algorithm and training framework.
\subsubsection{Dialogue Management Algorithm}
The model aims to predict the optimal dialogue act $a_t$ to maximize a reward function $\mathcal{R}(s_t,a_t)$. The system comprises a master policy $\theta$ and a set of $N$ sub-policies $\psi_0,…,\psi_N$. The master policy, with a discrete action space of size $N$, determines the appropriate sub-policy to use for the next $H$ turns, while each sub-policy $\psi_i$ manages the dialogue acts of the corresponding phase within an action space of size $N_{da}=13$. At each time step $t$, the algorithm operates as follows. If $t \mod H=0$, the master policy selects the next sub-policy: $A_ t =\theta(s_t^{master})$. Otherwise, the previous master action is reused: $A_t=A_{t-1}$. The sub-policy corresponding to $A_t$ then generates the next dialogue act: $a_t =\psi_{A_t}(s_t)$. This action $a_t$ influences the environment, producing a user response and updating the state to $s_{t+1}$ and $s_{t+1}^{master}$. The detailed algorithm is presented in Algorithm \ref{alg:dialog_management}

\subsubsection{Training Framework}  
The training framework leverages a model-based reinforcement learning (RL) approach. A model-based approach enables efficient reuse of dialogue turns across multiple iterations as policies evolve. Specifically, we utilize the Soft Actor-Critic (SAC) algorithm \cite{haarnoja2018soft}, which enhances the system's adaptability to new human users in online interactions. This approach allows for policy updates at each turn, maintaining the information from previous turns. Each training epoch targets a specific user profile and begins with cloning the master policy \( \theta \). The optimization process occurs in two phases. In the first phase, the master policy \( \theta \) is fixed, and the sub-policies \( \psi_0, \ldots, \psi_n \) are optimized using SAC. In the second phase, the sub-policies remain fixed while the cloned master policy \( \theta_{\text{clone}} \) is optimized using SAC. After these optimizations, the updated policies are evaluated. Finally, the master policy \( \theta \) is updated using the MAML algorithm. The complete training process is detailed in Algorithm \ref{alg:training_process}.
\begin{algorithm*}
\caption{Hierarchical Dialogue Management Algorithm}
\label{alg:dialog_management}
\begin{algorithmic}[1]
\State \textbf{Input:} State $s_t$ and $s^{master}_t$, Master policy $\theta$, Sub-policies $\psi_0, \ldots, \psi_n$, Time horizon $H$
\State \textbf{Initialize:} Master policy $\theta$, Sub-policies $\psi_0, \ldots, \psi_n$
\For{each time step $t$}
    \If{$t \mod H = 0$} 
        \State Compute master action $A_t = \theta(s^{master}_t)$
    \Else
        \State Reuse previous master action $A_t = A_{t-1}$
    \EndIf
    \State Select sub-policy $\psi_{A_t}$ based on $A_t$
    \State Compute dialogue act $a_t = \psi_{A_t}(s_t)$
    \State Apply action $a_t$ to the environment
    \State Observe user response and new state $s_{t+1}$ and  $s^{master}_{t+1}$
\EndFor
\end{algorithmic}
\end{algorithm*}
\begin{algorithm*}
\caption{Training Process for Hierarchical Dialogue Manager}
\label{alg:training_process}
\begin{algorithmic}[1]
\State \textbf{Input:} Master policy $\theta$, Sub-policies $\psi_0, \ldots, \psi_n$, User simulator, Replay buffer $D_{sub}$, Replay buffer $D_{master}$
\State \textbf{Initialize:} Master policy $\theta$, Sub-policies $\psi_0, \ldots, \psi_n$, Replay buffer $D_{sub}$, Replay buffer $D_{master}$
\For{each training epoch}
    \State Sample user profile $p$ and apply to the simulator
    \State \textbf{Phase 1: Sub-policy Optimization}
    \State Fix $\theta$ and for $N_{\text{sub}}$ dialogues:
    \For{each dialogue turn $t$}
        \State Generate transition \( (s_t, a_t, r_t, s_{t+1}, A_{t}) \) and store in replay buffer $D_{sub}$
        \State Sample a batch \( B \) from $D_{sub}$
        \For{each sub-policy $\psi_i$}
            \State Optimize $\psi_i$: \( \psi_i \gets \text{SAC}(\psi_i, B_{A_t=i}) \)
        \EndFor
    \EndFor
    \State \textbf{Phase 2: Master Policy Optimization}
    \State Clone master policy: $\theta_{\text{clone}} \gets \theta$
    \State Fix $\psi_0, \ldots, \psi_n$ and for $N_{\text{master}}$ dialogues:
    \For{each dialogue}
        \State Generate transition \( (s_t, a_t, r_t, s_{t+1}) \) and store in replay buffer $D_{master}$
        \State Optimize $\theta_{\text{clone}}$: \( \theta_{\text{clone}} \gets \text{SAC}(\theta_{\text{clone}}) \)
    \EndFor
    \State \textbf{Evaluation}
    \State Evaluate the updated policies $\theta_{\text{clone}}$ and $\psi_0, \ldots, \psi_n$ on the task
    \State \textbf{Master Policy Update}
    \State Update the master policy $\theta$ using the MAML algorithm: $\theta \gets \text{MAML}(\theta_{\text{clone}})$
    \State Empty  Replay buffer $D_{sub}$ and Replay buffer $D_{master}$
\EndFor
\end{algorithmic}
\end{algorithm*}

\section{Evaluation Environment}  
\label{sec:evaluation}  
In this section, we present the evaluation environment of our framework on a MI dialogue environment and compare our model with a state-of-the-art LLM baseline. We also perform ablations to demonstrate the efficiency of each of the model components, showing that incorporating knowledge on the flow of the dialogue in the dialogue manager development improves the resulting conversations.
%The code used for training and evaluation is available at \href{[ADD GITHUB LINK]}{[Add anonymous Github Link]}.  
\subsection{Baseline}
We use as a baseline a Nemo Instruct LLM\footnote{\label{footnote:nemo}\href{https://huggingface.co/mistralai/Mistral-Nemo-Instruct-2407}{https://huggingface.co/mistralai/Mistral-Nemo-Instruct-2407}} prompted as validated in \cite{steenstra2024virtual}. The prompt incorporates information on MI strategies, as well as techniques for addressing specific problems, such as \textit{Drinking}, \textit{Smoking}, and \textit{Sedentary Lifestyle}. This approach was validated in \cite{steenstra2024virtual} through experiments with human participants.
\subsection{Evaluation Environment}  
The evaluation environment includes a simulated user described in Section \ref{sec:user} with a specific profile \( P \) and a topic \( T \), where $ T \in \{ \text{Smoking, Alcohol, Sedentary Lifestyle} \} $. Additionally, the environment incorporates a Mistral LLM, specifically a Nemo Instruct LLM\footref{footnote:nemo}, which is prompted to generate both therapist and patient utterances based on the context, discussion theme, and dialogue act. This generation was validated in \cite{galland2024generating}. %An Utterance Encoder LLM: Nomic embed text\footnote{\href{https://huggingface.co/nomic-ai/nomic-embed-text-v1.5}{https://huggingface.co/nomic-ai/nomic-embed-text-v1.5}} is used to encode the utterances into a latent space representation.

\subsubsection{Simulate patients in MI}
\label{sec:user}
Simulating patients in MI has been explored in prior research. For instance, \cite{galland2024generating} proposed a prompt to simulate a user with a large language model (LLM) \cite{galland2024simulating}. This approach has been validated to produce contextually relevant, natural dialogue acts and utterances \cite{galland2024simulating,galland2024generating}, although the differences between user profiles have not been tested and might be a limitation of this simulator. We use this user simulator to train and evaluate our dialogue manager. In the following of this article, the term user refers to this user simulator.
\subsubsection{Action Space}
The agent operates in a discrete action space consisting of 13 possible dialogue acts, which are categorized into task-oriented dialogue acts and socially oriented dialogue acts. Task-oriented dialogue acts include \textit{Asking for Consent or Validation, Providing Medical Education and Guidance, Planning with the Patient, Giving a Solution, Asking about Current Emotions, Inviting a Shift in Outlook, Asking for Information}, and \textit{Reflection}. Socially-oriented dialogue acts include \textit{Empathic reactions, Acknowledging Progress and Encouraging, Backchanneling, Greeting or Closing, and Normalizing Experiences while Providing Reassurance}. This taxonomy was introduced in \cite{galland2024emmiannonymous}.

\subsubsection{State Space}
The agent's state space includes information from the most recent agent's and user's dialogue acts. User can use 9 different dialogue acts: \textit{Changing Unhealthy Behavior, Sustaining Unhealthy Behavior, Sharing Negative/Positive Feelings or Emotions, Sharing Personal Information, Realization or Understanding, Greeting or Closing, Backchanneling,} and \textit{Asking for Medical Information}. Additionally, the state space incorporates the current timestamp and an encoded representation of the dialogue context, which comprises the last three utterances.

\subsubsection{Master State Space}
The master policy's state space is composed of an approximation of Context knowledge, Engagement approximation, and Evocation approximation. Context knowledge approximation is measured by the number of times the user employs the \textit{Sharing Personal Information} dialogue act. Engagement approximation is determined by the number of times the user utilizes the \textit{Sharing Positive/Negative Feelings} dialogue act. Evocation approximation is quantified by the number of uses of the \textit{Understanding or New Perspective} dialogue act. 

\subsubsection{Reward Function}

The reward function is designed to predict therapy outcomes by assigning specific values to different user dialogue acts. Prior research underscores the critical role of user responses, such as \textit{sustain talk}, which is linked to poorer treatment outcomes \cite{magill2014technical}, and \textit{change talk}, which is associated with reduced risk behaviors during follow-up assessments \cite{magill2018meta}. Additionally, the reward function incentivizes structured progression through the MI phases. A reward of \( +5 \) is assigned for \textit{Changing Unhealthy Behavior}, as this represents the desired outcome, whereas a penalty of \( -5 \) is given for \textit{Sustaining Unhealthy Behavior}, which should be discouraged. In the Engagement phase, a reward of \( +50 \) is granted for expressing feelings. Once at least two emotions have been expressed by the user, a reward of \( +100 \) is assigned for providing information in the Focusing phase. After at least two pieces of information have been shared, a reward of \( +150 \) is given for evoking-related dialogue acts, culminating in a reward of \( +200 \) for planning-related dialogue acts.

\subsubsection{Episode Termination}
An episode concludes after 40 turns or when the agent performs a closing action, marking the end of the dialogue interaction.

\subsection{Hyperparameters}
The model is trained for 41 epochs, with 455 conversations conducted per epoch. Of these, 300 are used to train the master policy, while 150 are dedicated to training the sub-policies, and 5 to evaluation. To speed up the training process, 5 conversations are performed in parallel. There are 6 sub-policies in total, and the master policy is executed every 3 turns. The user's profile is fixed randomly at the beginning of each epoch. The sub-policies are trained with a learning rate of 
$10^{-4}$ and batch size of $1000$, the master policy uses a learning rate of $10^{-3}$ and batch size of $1000$,  and the MAML (Model-Agnostic Meta-Learning) algorithm operates with a learning rate of $4*10^{-4}$. Each network is composed of 2 linear layers intercalated with Leaky ReLU activation functions and a hidden size of $32$. The model is trained over 16 hours using a 42GB GPU.

\section{Results}

 \label{sec:results}
\begin{figure}
     \centering
     \includegraphics[width=0.7\linewidth]{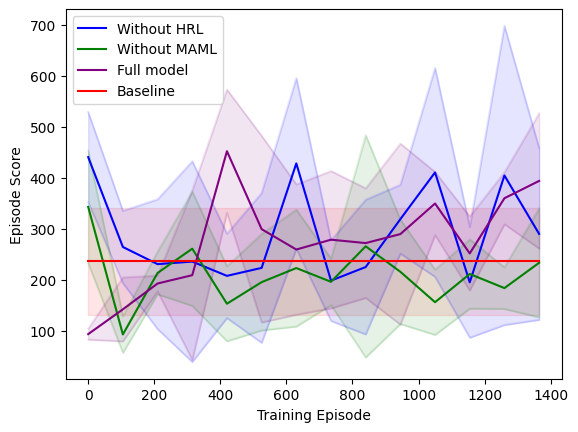}
\caption{Evolution of the reward function during training}
     \label{fig:reward}
\end{figure}

\begin{table}
\centering
\caption{Experiment Results with Mean Rewards (* = Statistical difference with baseline)}
\label{tab:experiment_results}
\resizebox{\linewidth}{!}{%
\begin{tabular}{lc}
\hline
\textbf{Experiment} & \textbf{Mean Reward (± SD)} \\
 & (. = difference with the baseline p<.1)\\
\hline
Baseline        & 235± 106                                 \\
\hline
\hline
Without MAML        & 233 ±  106                     \\
Without HRL         & 290 ± 169
\\
Full model  & \textbf{394 ± 132} (.)        \\
\hline
\end{tabular}
}
\end{table}
In this section, we present our results and ablation studies. 
Figure \ref{fig:reward} illustrates the mean reward evolution across all three user profiles—Open to Change, Resistant to Change, and Hesitant—throughout the training process. 
At each evaluation epoch, five conversations are conducted with each user profile. Additionally, Table \ref{tab:experiment_results} presents the final experimental results. Our model’s reward performance surpasses that of the baseline, demonstrating that conditioning an LLM with our dialogue manager enhances the proportion of desirable dialogue acts. We observe high variability in reward values throughout the training process. This fluctuation is primarily due to the inherent stochasticity of dialogue environments, where the same agent action in a given state can lead to different user responses. Such variability introduces additional challenges to training, making consistent improvement more difficult to achieve. Moreover, due to computational constraints, five dialogues at each evaluation epoch are conducted. Such a low number cannot represent the variability of potential dialogues, thus further contributing to the observed variance.
Despite this variability, an upward trend in performance is observed over the course of training. The RL model consistently outperforms the baseline in terms of average reward, demonstrating its effectiveness in promoting desirable dialogue behaviors.
\subsection{Ablation Studies}
We conduct two ablation studies to evaluate the impact of each design choice. In the first ablation study, we perform the same training procedure without employing MAML to train the master policy. In the second ablation study, we remove the hierarchical reinforcement learning (HRL) framework and train solely with the SAC algorithm (see Figure \ref{fig:reward} and 
Table \ref{tab:experiment_results}).

\paragraph{Effect of MAML}
The inclusion of MAML improves the accumulated reward (see Table \ref{tab:experiment_results}), suggesting that it enhances the learning of the master policy by explicitly accounting for variations across user profiles. Standard training can be biased by the sequence in which different user profiles are encountered, whereas MAML mitigates this by guiding the master policy toward an initialization that enables rapid adaptation to diverse users. 
\paragraph{Effect of HRL}
The effectiveness of HRL is further supported by the experimental results. Training with only the SAC algorithm leads to a lower accumulated reward, likely because the phase-based structure of MI dialogues is more challenging to capture without hierarchical modeling.

\section{Interpretation}
\label{sec:interpretation}

In this section, we analyze the generated dialogues and examine how our design choices influence them. We investigate the different MI phases to determine whether they emerge as expected and assess the impact of HRL. Additionally, we explore variations across user profiles and evaluate the effect of MAML on the generated dialogues.

\paragraph{Differences Between Phases}
\begin{figure}[t!]
    \centering
    \includegraphics[width=\linewidth]{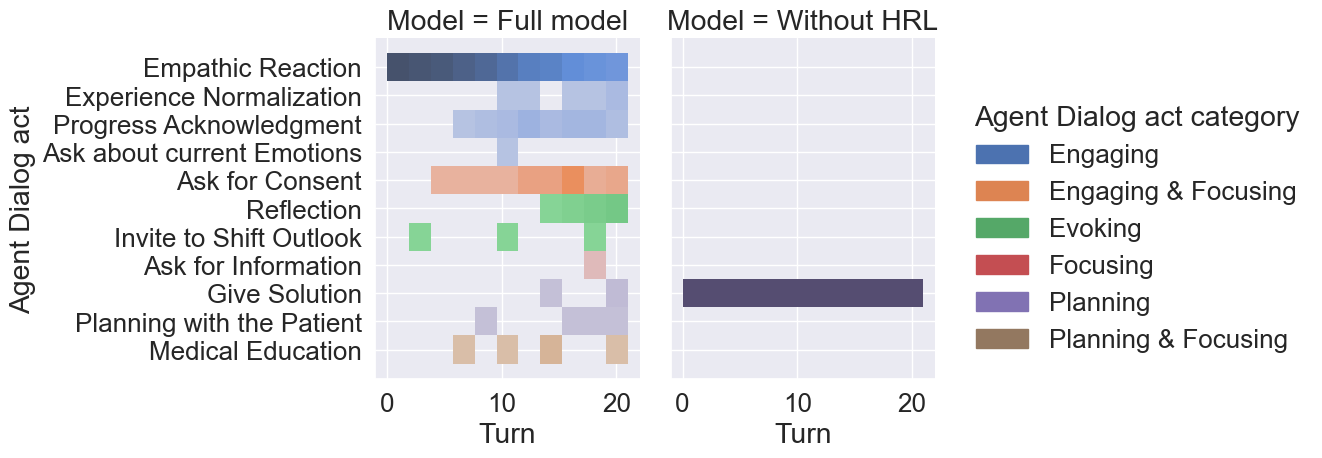}
    \caption{Dialogue act distribution over time, highlighting different dialogue phases for the full model. \small{The intensity of the color is proportional to the use of the corresponding dialogue act in the turn.}}
    \label{fig:interpphase}
\end{figure}
To analyze the MI phases, we examine the distribution of dialogue acts across different dialogue turns. Dialogue acts associated with the Engaging phase, such as asking about emotions or sharing emotions, should be more prevalent at the beginning of the conversation, whereas those related to the Planning phase, such as providing solutions or promoting behavior change, should appear more frequently towards the end \cite{miller2012motivational}. While Engaging should occur throughout the entire dialogue, the later stages should be more focused on Planning. The phases are interwoven rather than strictly sequential %(see Figure \ref{fig:mi_phases}). 
Figure \ref{fig:interpphase} shows that the full model employs engagement-related dialogue acts throughout the conversation and uses focusing-related actions in the last three quarters of the interaction. It also uses evoking- and planning-related action towards the end of the interaction. However, the HRL ablation only uses a singular action: "Give solution". This strategy may provide immediate rewards with the simulated users, but does not adhere to the MI principle, making the full model with HRL more diverse and MI compliant.

\paragraph{Differences between user profiles}
\begin{figure}
\centering
\includegraphics[width=\linewidth]{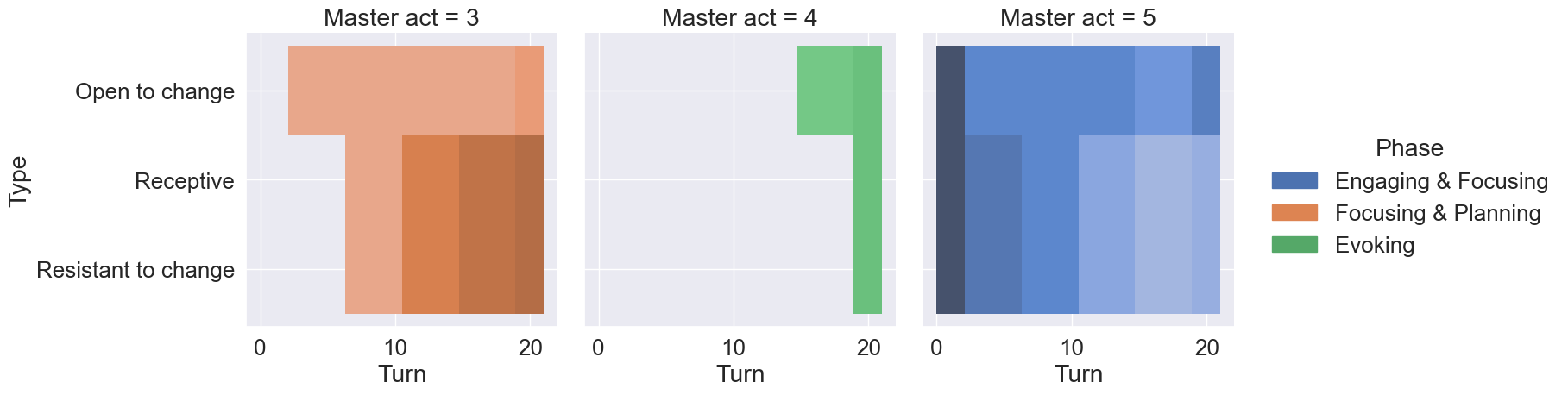}
\caption{Distribution of phase activation over time across different user profiles. \small{The intensity of the color is proportional to the use of the phase in the turn.}}
\label{fig:interp_user}
\end{figure}

In this section, we examine the impact of meta-learning on training the master policy. Figure \ref{fig:interp_user} illustrates the distribution of master actions activation over time for different user profiles for the full model. For the model trained without meta-learning, we observe a collapse of the master policy to a single dominant action across interactions. This highlights the difficulty of learning a generalized policy that performs well across diverse user profiles without explicit mechanisms for adaptation. In contrast, the use of meta-learning allows the master policy to maintain variability and adaptability in its actions. In Figure \ref{fig:interp_user}, we associate each master action with MI phases based on the predominant dialogue acts typically observed in those phases and master actions, providing interpretability into the policy's learned structure. The models effectively differentiate between distinct phases, initially engaging in Engaging/Focusing phases (Master action 5) before transitioning into Focusing/Planning phases (Master action 3) and Evoking phase (Master action 4).The engaging phase is quicker for Open to change users, as they do human therapist, aligning with prior findings \cite{galland2024emmiannonymous}. Indeed, open-to-change patients require less rapport-building with the therapist, as they choose to initiate the conversation. This qualitative analysis highlights the benefits of incorporating meta-learning, as it enhances the model’s ability to structure MI phases effectively and provide a more personalized dialogue experience.

\section{Conclusion}

In this paper, we present a dialogue manager for MI dialogue design, addressing the unique challenges posed by this profile of interaction. We leverage HRL to model the structured phases of MI and employ meta-learning to enhance adaptability across diverse user profiles. Our findings demonstrate that the proposed dialogue manager outperforms an LLM baseline in terms of reward. Additionally, our analysis of the generated conversations provides valuable insights into how HRL and meta-learning contribute to the structured yet adaptive nature of the dialogue.

% Ne compte pas dans la limite de pages
\section{Limitations}
The current framework is trained using a single implementation of a simulated user, which limits its generalizability. To fully assess its effectiveness, the model is currently being tested with human participants to capture a broader range of user behaviors and characteristics.

Moreover, the simulator is based on a Mistral LLM, which requires significant processing time to generate user behaviors. This limitation constrains the training capacity, as complex reinforcement learning (RL) problems like this one require extensive trial and error. As a result, certain design choices—such as small batch sizes and a limited number of conversations per epoch—were necessary, contributing to the observed training instability. Addressing this issue in future work could lead to more robust and efficient training.

Additionally, the analysis of dialogue phases currently relies on predefined heuristics, making it inherently subjective. A more rigorous approach would involve annotation and validation by professional MI annotators to ensure alignment with clinical practices, thereby improving the system’s reliability.

\section{Ethical Implications}
This work introduces a dialogue manager designed for Motivational Interviewing interactions. Its objective is not to replace therapists but to provide supplementary support or serve as an introduction to therapy. The focus of this research is exclusively on the dialogue management component, which operates within a constrained set of possible actions. All the LLMs are run locally, and no sensitive information is sent to outside services.

Given the sensitive nature of such applications, careful examination and validation of the language model outputs remain essential. It is imperative to emphasize in both the codebase and accompanying documentation that these interactions are not intended to replace professional therapists but rather to complement their efforts in appropriate contexts. Ensuring transparency and adherence to ethical standards is fundamental to responsibly deploying this technology. Additionally, it would be important to validate the approach with MI practitioners before using it in any real world context.

\section{Acknowledgement}
This work was partially funded by the ANR-DFG-JST Panorama and ANR-JST-CREST TAPAS (19-JSTS-0001-01) projects
\bibliography{acl_latex}

\begin{thebibliography}{29}
\providecommand{\natexlab}[1]{#1}

\bibitem[{Abu-Rasheed et~al.(2024)Abu-Rasheed, Abdulsalam, Weber, and Fathi}]{abu2024supporting}
Hasan Abu-Rasheed, Mohamad~Hussam Abdulsalam, Christian Weber, and Madjid Fathi. 2024.
\newblock Supporting student decisions on learning recommendations: An llm-based chatbot with knowledge graph contextualization for conversational explainability and mentoring.
\newblock \emph{arXiv preprint arXiv:2401.08517}.

\bibitem[{anonymous(2024)}]{galland2024emmiannonymous}
anonymous. 2024.
\newblock anonymous.

\bibitem[{Baktash and Dawodi(2023)}]{baktash2023gpt}
Jawid~Ahmad Baktash and Mursal Dawodi. 2023.
\newblock Gpt-4: A review on advancements and opportunities in natural language processing.
\newblock \emph{arXiv preprint arXiv:2305.03195}.

\bibitem[{Cameron et~al.(2017)Cameron, Cameron, Megaw, Bond, Mulvenna, O’Neill, Armour, and McTear}]{cameron2017towards}
Gillian Cameron, David Cameron, Gavin Megaw, Raymond Bond, Maurice Mulvenna, Siobhan O’Neill, Cherie Armour, and Michael McTear. 2017.
\newblock Towards a chatbot for digital counselling.
\newblock In \emph{Proceedings of the 31st International BCS Human Computer Interaction Conference (HCI 2017) 31}, pages 1--7.

\bibitem[{Denecke et~al.(2020)Denecke, Vaaheesan, and Arulnathan}]{denecke2020mental}
Kerstin Denecke, Sayan Vaaheesan, and Aaganya Arulnathan. 2020.
\newblock A mental health chatbot for regulating emotions (sermo)-concept and usability test.
\newblock \emph{IEEE Transactions on Emerging Topics in Computing}, 9(3):1170--1182.

\bibitem[{Finn et~al.(2017)Finn, Abbeel, and Levine}]{finn2017model}
Chelsea Finn, Pieter Abbeel, and Sergey Levine. 2017.
\newblock Model-agnostic meta-learning for fast adaptation of deep networks.
\newblock In \emph{International conference on machine learning}, pages 1126--1135. PMLR.

\bibitem[{Fiske et~al.(2019)Fiske, Henningsen, and Buyx}]{fiske2019your}
Amelia Fiske, Peter Henningsen, and Alena Buyx. 2019.
\newblock Your robot therapist will see you now: ethical implications of embodied artificial intelligence in psychiatry, psychology, and psychotherapy.
\newblock \emph{Journal of medical Internet research}, 21(5):e13216.

\bibitem[{Galland et~al.(2024{\natexlab{a}})Galland, Pelachaud, and Pecune}]{galland2024generating}
Lucie Galland, Catherine Pelachaud, and Florian Pecune. 2024{\natexlab{a}}.
\newblock Generating unexpected yet relevant user dialog acts.
\newblock In \emph{Proceedings of the 25th Annual Meeting of the Special Interest Group on Discourse and Dialogue}, pages 192--203.

\bibitem[{Galland et~al.(2024{\natexlab{b}})Galland, Pelachaud, and Pecune}]{galland2024simulating}
Lucie Galland, Catherine Pelachaud, and Florian Pecune. 2024{\natexlab{b}}.
\newblock Simulating patient oral dialogues: A study on naturalness and %coherence of conditioned large language models.
\newblock In \emph{Proceedings of the ACM International Conference on Intelligent Virtual Agents}, pages 1--4.

\bibitem[{Haarnoja et~al.(2018)Haarnoja, Zhou, Abbeel, and Levine}]{haarnoja2018soft}
Tuomas Haarnoja, Aurick Zhou, Pieter Abbeel, and Sergey Levine. 2018.
\newblock Soft actor-critic: Off-policy maximum entropy deep reinforcement learning with a stochastic actor.
\newblock In \emph{International conference on machine learning}, pages 1861--1870. PMLR.

\bibitem[{Hadi et~al.(2024)Hadi, Al~Tashi, Shah, Qureshi, Muneer, Irfan, Zafar, Shaikh, Akhtar, Wu et~al.}]{hadi2024large}
Muhammad~Usman Hadi, Qasem Al~Tashi, Abbas Shah, Rizwan Qureshi, Amgad Muneer, Muhammad Irfan, Anas Zafar, Muhammad~Bilal Shaikh, Naveed Akhtar, Jia Wu, et~al. 2024.
\newblock Large language models: a comprehensive survey of its applications, challenges, limitations, and future prospects.
\newblock \emph{Authorea Preprints}.

\bibitem[{He et~al.(2024)He, Liao, Cao, Liu, Liu, Chen, and Qin}]{he2024planning}
Tao He, Lizi Liao, Yixin Cao, Yuanxing Liu, Ming Liu, Zerui Chen, and Bing Qin. 2024.
\newblock Planning like human: A dual-process framework for dialogue planning.
\newblock \emph{arXiv preprint arXiv:2406.05374}.

\bibitem[{Kanaoka and Mutlu(2015)}]{kanaoka2015designing}
Toshikazu Kanaoka and Bilge Mutlu. 2015.
\newblock Designing a motivational agent for behavior change in physical activity.
\newblock In \emph{Proceedings of the 33rd Annual ACM Conference Extended Abstracts on Human Factors in Computing Systems}, pages 1445--1450.

\bibitem[{Li et~al.(2016)Li, Monroe, Ritter, Galley, Gao, and Jurafsky}]{li2016deep}
Jiwei Li, Will Monroe, Alan Ritter, Michel Galley, Jianfeng Gao, and Dan Jurafsky. 2016.
\newblock Deep reinforcement learning for dialogue generation.
\newblock \emph{arXiv preprint arXiv:1606.01541}.

\bibitem[{Magill et~al.(2018)Magill, Apodaca, Borsari, Gaume, Hoadley, Gordon, Tonigan, and Moyers}]{magill2018meta}
Molly Magill, Timothy~R Apodaca, Brian Borsari, Jacques Gaume, Ariel Hoadley, Rebecca~EF Gordon, J~Scott Tonigan, and Theresa Moyers. 2018.
\newblock A meta-analysis of motivational interviewing process: Technical, relational, and conditional process models of change.
\newblock \emph{Journal of consulting and clinical psychology}, 86(2):140.

\bibitem[{Magill et~al.(2014)Magill, Gaume, Apodaca, Walthers, Mastroleo, Borsari, and Longabaugh}]{magill2014technical}
Molly Magill, Jacques Gaume, Timothy~R Apodaca, Justin Walthers, Nadine~R Mastroleo, Brian Borsari, and Richard Longabaugh. 2014.
\newblock The technical hypothesis of motivational interviewing: A meta-analysis of mi’s key causal model.
\newblock \emph{Journal of consulting and clinical psychology}, 82(6):973.

\bibitem[{Miller and Rollnick(2012)}]{miller2012motivational}
William~R Miller and Stephen Rollnick. 2012.
\newblock \emph{Motivational interviewing: Helping people change}.
\newblock Guilford press.

\bibitem[{Olafsson et~al.(2020)Olafsson, Wallace, and Bickmore}]{olafsson2020towards}
Stefan Olafsson, Byron~C Wallace, and Timothy~W Bickmore. 2020.
\newblock Towards a computational framework for automating substance use counseling with virtual agents.
\newblock In \emph{AAMAS}, volume~19, pages 9--13. Auckland.

\bibitem[{Pecune and Marsella(2020)}]{pecune2020framework}
Florian Pecune and Stacy Marsella. 2020.
\newblock A framework to co-optimize task and social dialogue policies using reinforcement learning.
\newblock In \emph{Proceedings of the 20th ACM International Conference on Intelligent Virtual Agents}, pages 1--8.

\bibitem[{Prochaska et~al.(2021)Prochaska, Vogel, Chieng, Kendra, Baiocchi, Pajarito, and Robinson}]{prochaska2021therapeutic}
Judith~J Prochaska, Erin~A Vogel, Amy Chieng, Matthew Kendra, Michael Baiocchi, Sarah Pajarito, and Athena Robinson. 2021.
\newblock A therapeutic relational agent for reducing problematic substance use (woebot): development and usability study.
\newblock \emph{Journal of medical Internet research}, 23(3):e24850.

\bibitem[{Shidara et~al.(2020)Shidara, Tanaka, Adachi, Kanayama, Sakagami, Kudo, and Nakamura}]{shidara2020analysis}
Kazuhiro Shidara, Hiroki Tanaka, Hiroyoshi Adachi, Daisuke Kanayama, Yukako Sakagami, Takashi Kudo, and Satoshi Nakamura. 2020.
\newblock Analysis of mood changes and facial expressions during cognitive behavior therapy through a virtual agent.
\newblock In \emph{Companion Publication of the 2020 International Conference on Multimodal Interaction}, pages 477--481.

\bibitem[{Steenstra et~al.(2024)Steenstra, Nouraei, Arjmand, and Bickmore}]{steenstra2024virtual}
Ian Steenstra, Farnaz Nouraei, Mehdi Arjmand, and Timothy Bickmore. 2024.
\newblock Virtual agents for alcohol use counseling: Exploring llm-powered motivational interviewing.
\newblock In \emph{Proceedings of the 24th ACM International Conference on Intelligent Virtual Agents}, pages 1--10.

\bibitem[{Takanobu et~al.(2020)Takanobu, Liang, and Huang}]{takanobu2020multi}
Ryuichi Takanobu, Runze Liang, and Minlie Huang. 2020.
\newblock Multi-agent task-oriented dialog policy learning with role-aware reward decomposition.
\newblock \emph{arXiv preprint arXiv:2004.03809}.

\bibitem[{Walker(2000)}]{walker2000application}
Marilyn~A Walker. 2000.
\newblock An application of reinforcement learning to dialogue strategy selection in a spoken dialogue system for email.
\newblock \emph{Journal of Artificial Intelligence Research}, 12:387--416.

\bibitem[{Weber et~al.(2018)Weber, Ritschel, Aslan, Lingenfelser, and Andr{\'e}}]{weber2018shape}
Klaus Weber, Hannes Ritschel, Ilhan Aslan, Florian Lingenfelser, and Elisabeth Andr{\'e}. 2018.
\newblock How to shape the humor of a robot-social behavior adaptation based on reinforcement learning.
\newblock In \emph{Proceedings of the 20th ACM international conference on multimodal interaction}, pages 154--162.

\bibitem[{Xu et~al.(2023)Xu, Xu, Wang, Liu, Zhu, and McAuley}]{xu2023small}
Canwen Xu, Yichong Xu, Shuohang Wang, Yang Liu, Chenguang Zhu, and Julian McAuley. 2023.
\newblock Small models are valuable plug-ins for large language models.
\newblock \emph{arXiv preprint arXiv:2305.08848}.

\bibitem[{Yao et~al.(2023)Yao, Heinecke, Niebles, Liu, Feng, Xue, Murthy, Chen, Zhang, Arpit et~al.}]{yao2023retroformer}
Weiran Yao, Shelby Heinecke, Juan~Carlos Niebles, Zhiwei Liu, Yihao Feng, Le~Xue, Rithesh Murthy, Zeyuan Chen, Jianguo Zhang, Devansh Arpit, et~al. 2023.
\newblock Retroformer: Retrospective large language agents with policy gradient optimization.
\newblock \emph{arXiv preprint arXiv:2308.02151}.

\bibitem[{Yu et~al.(2023)Yu, Chen, and Yu}]{yu2023prompt}
Xiao Yu, Maximillian Chen, and Zhou Yu. 2023.
\newblock Prompt-based monte-carlo tree search for goal-oriented dialogue policy planning.
\newblock \emph{arXiv preprint arXiv:2305.13660}.

\bibitem[{Zhang et~al.(2020)Zhang, Liao, Zhu, Chua, Liu, Huang, and Huang}]{zhang2020learning}
Zheng Zhang, Lizi Liao, Xiaoyan Zhu, Tat-Seng Chua, Zitao Liu, Yan Huang, and Minlie Huang. 2020.
\newblock Learning goal-oriented dialogue policy with opposite agent awareness.
\newblock \emph{arXiv preprint arXiv:2004.09731}.

\end{thebibliography}

% \appendix

% \section{Example Appendix}
% \label{sec:appendix}

% This is an appendix.

\end{document}